\pdfoutput=1
\documentclass[11pt]{article}
\usepackage[]{acl}
\usepackage{times}
\usepackage{latexsym}
\usepackage{graphicx}
\usepackage{amssymb}
\usepackage[T1]{fontenc}
\usepackage[utf8]{inputenc}
\usepackage{microtype}
\usepackage{amsmath}
\usepackage{amsmath}
\usepackage{amssymb}
\usepackage{mathtools}
\usepackage{amsthm}
\usepackage{mdframed} 
\usepackage[capitalize,noabbrev]{cleveref}
\usepackage{booktabs} 
\usepackage{multirow} 
\usepackage{makecell} 
\usepackage{xcolor} 
\usepackage{colortbl} 
\usepackage{enumitem}
\usepackage{ragged2e}

\title{MIH-TCCT: Mitigating Inconsistent Hallucinations in LLMs via Event-Driven Text-Code Cyclic Training}

\author{
\normalfont 
Xinxin You\textsuperscript{$1$,†}, Xien Liu\textsuperscript{$1$,†}, Qixin Sun\textsuperscript{$2$,†}, Huan Zhang\textsuperscript{$3$}, Kaiyin Zhou\textsuperscript{$4$}, \\
Shaohui Liu\textsuperscript{$4$}, Guoping Hu\textsuperscript{$3$}, Shijin Wang\textsuperscript{$3$}, Si Liu\textsuperscript{$2$}, Ji Wu\textsuperscript{$1$} \\[0.3em]
{\normalsize $^1$Tsinghua University, Beijing, China \quad $^2$Beihang University, Beijing, China} \\[0.1em]
{\normalsize $^3$iFLYTEK Research, Beijing, China \quad $^4$Beijing University of Posts and Telecommunications, Beijing, China} \\[0.1em]
{\normalsize \texttt{yxx23@mails.tsinghua.edu.cn}}
}
\date{}

\begin{document}
\maketitle




\begin{abstract}

Recent methodologies utilizing synthetic datasets have aimed to address inconsistent hallucinations in large language models (LLMs); however, these approaches are primarily tailored to specific tasks, limiting their generalizability. Inspired by the strong performance of code-trained models in logic-intensive domains, we propose a novel framework that leverages event-based text to generate corresponding code and employs cyclic training to transfer the logical consistency of code to natural language effectively. Our method significantly reduces inconsistent hallucinations across three leading LLMs and two categories of natural language tasks while maintaining overall performance. This framework effectively alleviates hallucinations without necessitating adaptation to downstream tasks, demonstrating generality and providing new perspectives to tackle the challenge of inconsistent hallucinations.

\end{abstract}

\section{Introduction}

\begin{figure}[t]
    \includegraphics[height=6.5cm,width=7.8cm]{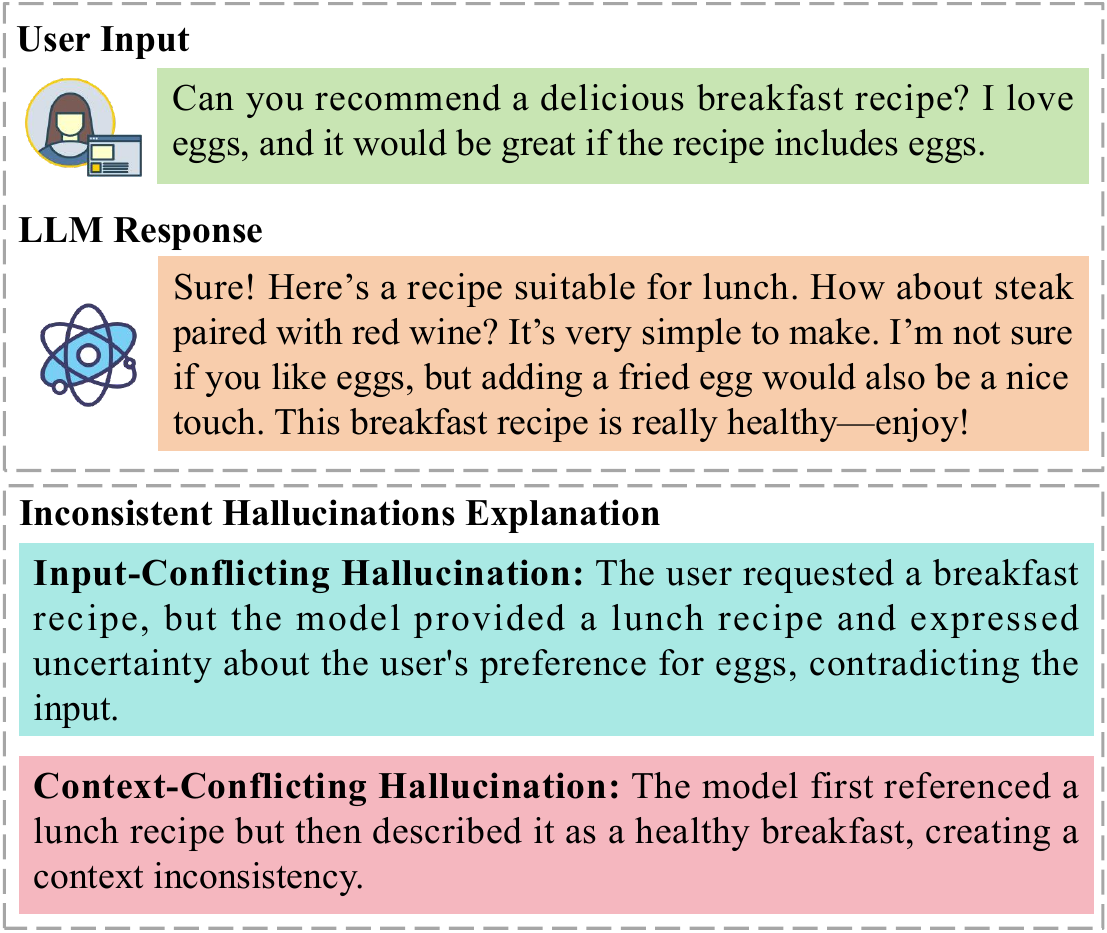}
    \centering
    \caption{Two types of inconsistent hallucinations occurred in LLM responses.}
    \label{fig:inconsistent hallucinations}
    \end{figure}

Achieving human-like logical consistency in reasoning is essential for advancing artificial general intelligence (AGI) \cite{achiam2023gpt,pan2023logic}. Despite significant advancements in large language models (LLMs) in addressing a variety of natural language processing tasks \cite{jiang2024survey,mundlerself,ribeirostreet}, these models still tend to produce hallucinations due to their inability to maintain logical rigor \cite{zhang2023siren,bao2024faithbench}. These manifest in two main forms: 1) Input-conflicting hallucination, where LLMs generate content that diverges from the provided input \cite{dale2023detecting,zhong2021qmsum}; and 2) Context-conflicting hallucination, where LLMs create content that contradicts information they have previously generated \cite{pu2023summarization,liu2024lost}, as shown in Figure \ref{fig:inconsistent hallucinations}. Given that these hallucinations may occur in various domains, there is an urgent need for a universal approach to effectively address these issues across different tasks. 

Recent studies have emphasized the effectiveness of synthetic datasets in mitigating in-consistent hallucinations related to specific tasks. For instance, synthetic mathematical datasets \cite{toshniwal2024openmathinstruct,huang2024key} have shown reduced contradictions in the calculation process and significant improvements in solving mathematical problems. Additionally, some research has synthesized reasoning data in the logical reasoning \cite{nie2020adversarial,saeed2021rulebert} field while providing the reasoning process \cite{dalvi2021explaining} during synthesis. LOGIC-LM and SymbCOT \cite{pan2023logic,xu2024faithful} integrate LLMs with symbolic solvers to enhance logical problem-solving. Unfortunately, the synthetic datasets that have been developed are often tailored to specific tasks or domains, which can restrict their generalizability.



Advancements in training LLMs on extensive code datasets, such as CoCoGen and CodeRL, have significantly reduced inconsistent hallucinations in code generation tasks \cite{madaan2022language,le2022coderl}. However, these improvements have not effectively transferred to general natural language tasks due to fundamental differences in semantics, language styles, and structures between code and natural language. The precision of code contrasts with the ambiguity of human language, posing challenges in adapting code-focused techniques for natural language understanding and generation. Recent approaches, such as program-of-thought \cite{chen2023program} and program-assisted LMs \cite{gao2023pal}, have sought to bridge this divide by interpreting natural language problems and generating corresponding code solutions. However, these methodologies fail to successfully transfer the logical consistency capabilities gained from code training to a broader spectrum of NLP tasks.

\begin{figure}[t]
    \includegraphics[height=4.5cm,width=7.8cm]{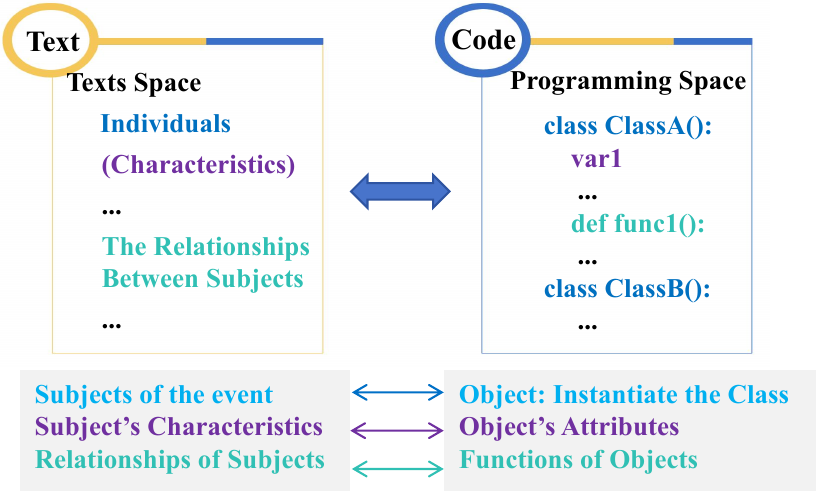}
    \centering
    \caption{The Correspondence Between Event-Driven 
Text and Programming Language.}
    \label{fig:compare}
    \end{figure}

Inspired by code-oriented LLMs that excel in generating logically consistent code, we propose that integrating code data is vital for enhancing the ability of LLMs to maintain logical consistency across various NLP applications. Furthermore, we observe that there exists a correspondence between the structure of event-based text and that of code. The event-based text can be transformed into structured code, utilizing programming constructs such as classes, objects, and functions, with the correspondence between the two structures shown in Figure \ref{fig:compare}. Thus, we introduce MIH-TCCT: Mitigating Inconsistent Hallucinations in LLMs via Event-Driven Text-Code Cyclic Training. This method facilitates the cyclic training of code and the original text, continuously aligning the stylistic characteristics of the two "languages" to improve dual generation capabilities. Ultimately, this approach aims to transfer the logical rigor of code data into natural language, fundamentally enhancing the logical consistency of LLM outputs. This novel integration not only fosters more reliable reasoning processes, but also paves the way for developing AI systems capable of delivering coherent and contextually accurate responses, representing a significant advancement in the capabilities of language models.

We summarize the contributions of our method as follows:
\begin{itemize}[left=0pt]
    \item We propose MIH-TCCT, a novel framework that enhances the logical consistency of LLMs outputs across diverse NLP tasks by cyclically generating event-based text and corresponding code. This framework effectively transfers the logical rigor of code to natural language, addressing critical challenges related to inconsistent hallucination.
\item We identify a clear correspondence between the structures of event-based text and code. This finding not only supports our cyclic training methodology but also paves the way for future research bridging the gap between textual and coding modalities.
\item Through our method, LLMs achieve the capability to maintain logical consistency, effectively mitigating hallucinations without adapting to downstream tasks, thereby demonstrating generalizability. Experiments across three LLMs and two categories of natural language tasks reveal a significant reduction in hallucinations.
\end{itemize}

\section{Related Work}

\subsection{Synthetic Data to Reduce Hallucinations}

Recent investigations indicate that hallucination phenomena in large language models (LLMs) may stem from inherent issues within the training data. This highlights the necessity for high-quality reasoning datasets to mitigate these inconsistencies \cite{zhang2023siren,jiang2024survey}. Notable examples include LOGIQA \cite{liu2023logiqa}, a dataset of logical reasoning derived from the Chinese Civil Service Examination; RECLOR \cite{yureclor} and AR-LSAT \cite{zhong2021ar}, based on standardized graduate admissions exams; and FOLIO \cite{han2022folio}, a dataset featuring first-order logic annotations. However, the resource-intensive and time-consuming nature of their creation limits the scalability and accessibility of training applications.

As a result, researchers are increasingly exploring synthetic data as a viable alternative \cite{achiam2023gpt,dubey2024llama}. Significant synthetic mathematical datasets include KPDDS, OpenMathInstruct-1, and MathGenie \cite{toshniwal2024openmathinstruct,huang2024key,lu2024mathgenie}. KPDDS synthesizes question-answer pairs using key points and exemplars, yielding KPMath, which encompasses over one million pairs. OpenMathInstruct-1 comprises 1.8 million problem-solution pairs synthesized from code-interpreter solutions for GSM8K and MATH benchmarks via the Mixtral model. MathGenie enhances a seed dataset by generating new questions through a back-translation approach. Despite these advancements, synthetic datasets are often tailored to specific tasks, particularly mathematics, which may restrict their generalizability and performance across diverse real-world applications. This underscores the necessity for further research to broaden their applicability.

\subsection{COT Data to Reduce Hallucinations}

Based on the investigation of synthetic datasets, Chain-of-Thought (CoT) strategies have also reduced hallucinations during reasoning tasks \cite{wei2022chain}. Some research has synthesized reasoning data in the field of logical reasoning \cite{nie2020adversarial,saeed2021rulebert} while providing the reasoning process \cite{dalvi2021explaining} during synthesis. For example, LOGIC-LM integrates LLMs with symbolic solvers, transforming natural language problems into symbolic formulations to minimize inconsistencies \cite{pan2023logic}. Similarly, SymbCOT enhances CoT prompting by incorporating symbolic expressions and logical rules \cite{xu2024faithful}.

The expressiveness of symbolic solvers limits the applicability of these models. Not all problems can be easily encoded in first-order logic, and complexities may arise when dealing with intricate grammatical structures, such as those found in probabilistic soft logic. Therefore, while these approaches show promise, they have constraints in flexibility and generalizability, highlighting the need for solutions that can effectively address a broader range of reasoning scenarios.

\subsection{Code Data to Reduce Hallucinations}

Recent research has shown that training large language models (LLMs) on code datasets can reduce in-consistent hallucinations. Notable models such as CoCoGen and CodeRL have enhanced performance in structured reasoning and code generation by leveraging the strict syntax and semantics inherent in programming languages \cite{madaan2022language,le2022coderl}. However, the benefits observed in code-related tasks do not readily extend to general natural language processing (NLP) tasks due to the fundamental semantic differences between the two domains.

Innovative approaches like Program of Thought (PoT) utilize LLMs, particularly Codex, to express reasoning processes as programs while offloading computational tasks to external systems \cite{chenprogram}. Despite this novel framework, PoT may encounter challenges in contexts where translating natural language into code is not straightforward, which can introduce potential errors. Similarly, Program-Aided Language Models (PAL) enable LLMs to decompose natural language problems into executable steps, delegating execution to environments such as Python interpreters \cite{gao2023pal}. While this structural separation simplifies the model's role, it also restricts its ability to engage in comprehensive reasoning, making it susceptible to errors during decomposition. Consequently, the effective transfer of logical consistency gained from code training to broader NLP applications remains an open challenge, highlighting the need for further research.

\begin{figure*}[t]
    \includegraphics[height=3cm,width=16cm]{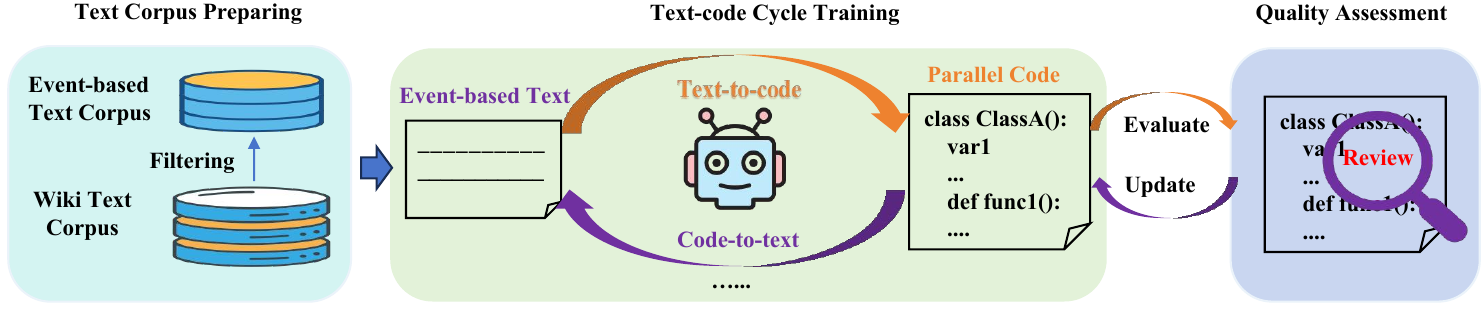}
    \centering
    \caption{An overview of our proposed framework begins with filtering event-based text, followed by cyclic generation training of event-based text and parallel code based on their transformation relationship. In each iteration, a quality evaluation module is employed to assess the quality of the generated parallel code until multiple iterations result in improved capabilities in parallel corpus generation, ultimately achieving alignment between the two corpora.}
    \label{fig:prompt2}
    \end{figure*}


\section{Method}

Our method is grounded in a novel insight: texts containing event descriptions reveal a coherent structural framework that can be seamlessly mapped to structured programming languages. These events—defined by participants, their attributes, causes, processes, and outcomes—form a rich structured information framework. This clear delineation allows us to effectively map these elements onto programming constructs such as functions, objects, and classes, as illustrated in Figure \ref{fig:compare}. For example, participants can be modeled as class instances, with their characteristics represented as attributes. Additionally, the causes and processes of events can be articulated through functions, enabling the translation of natural language descriptions into structured code. This cognitive framework bridges the gap between textual and coding modalities and supports our cyclic training methodology.

In the following sections, we will focus on several key study modules. Section 3.1 will define the problem and outline the research objectives. Section 3.2 will address the filtering of appropriate texts that describe events. In Section 3.3, we will discuss how the cyclic generation of text and code enhances the logical consistency of LLMs. Section 3.4 will present methodologies for assessing the quality of the generated code.

\subsection{Problem Definition}

This study investigates a large language model (LLM) that is tasked with generating text, represented as a sequence of sentences, denoted by:
\begin{equation}
\mathbf{x} = [x_1, x_2, \ldots, x_{|\mathbf{x}|}],
\end{equation}
where \( \mathbf{x} \) representing the entire sequence of generated sentences, and \( |\mathbf{x}| \) denotes the total number of sentences in that sequence. Each sentence is represented by \( x_i \).

The generation process is typically influenced by a user-defined prompt \( p \) that outlines specific objectives for the text generation and an input text \( t \) that provides relevant background information. The generation of the sequence of sentences \( \mathbf{x} \) can be expressed probabilistically as:
\begin{equation}
\mathbf{x} \sim LLM(\cdot | p, t),
\end{equation}
We investigate the phenomenon of context inconsistency in outputs generated by large language models (LLMs), specifically focusing on two types of inconsistencies.

\noindent \textbf{Input-Conflicting Hallucination}: We define input-conflicting hallucination as occurring when the generated sequence \( \mathbf{x} \) is inconsistent with either the prompt \( p \) or the text \( t \). This can be formally expressed as:
\begin{equation}
H_{incon}(\mathbf{x}, p) \quad \text{if} \quad (\mathbf{x} \neq p) \land \text{Contradict}(\mathbf{x}, p).
\end{equation}
\begin{equation}
H_{incon}(\mathbf{x}, t) \quad \text{if} \quad (\mathbf{x} \neq t) \land \text{Contradict}(\mathbf{x}, t).
\end{equation}
Using \( H_{incon}(\mathbf{x}, p) \) as an example, which indicates that a content conflict occurs when two conditions are met: \( \mathbf{x} \neq p \) (the two sentences are not identical) and \( \text{Contradict}(\mathbf{x}, p) \) (the sentences \( \mathbf{x} \) and \( p \) contradict each other, making them logically inconsistent).

\noindent \textbf{Context-Conflicting Hallucination}: Additionally, we define context-conflicting hallucination as the scenario where the generated output \( \mathbf{x} \) is composed of two parts, \( x_1 \) and \( x_2 \), generated under the same prompt \( p \) and context \( t \):
\begin{equation}
\mathbf{x} = [x_1, x_2] \sim LLM(\cdot | p, t).
\end{equation}
We assert that the pair \( (x_1, x_2) \) represents a context conflict when the two generated sentences are logically inconsistent. This can be expressed as:
\begin{equation}
H_{incon}(x_1, x_2) \quad \text{if} \quad (x_1 \neq x_2) \land \text{Contradict}(x_1, x_2).
\end{equation}

\subsection{Text Filtering for Implicit Events}

This section outlines the methodology employed to filter text containing event descriptions suitable for conversion into structured code. A crucial aspect of this process is establishing an evaluation mechanism, denoted as \( E \), which verifies whether the current text \( t \) meets the necessary criteria for transformation into code. To implement this filtering procedure, we utilize the GPT3.5-turbo model, guided byTo implement this filtering procedure, we employ the GPT-3.5 Turbo model, guided by the prompt detailed in Appendix A.1. The classification of a text \( t \) can be formally represented as follows:
\begin{equation}
D(t) = 
\begin{cases} 
1 & \text{if } E_{gpt}(t) = \text{True} \\ 
0 & \text{if } E_{gpt}(t) = \text{False} 
\end{cases}
\end{equation}
In this representation, \( D(t) = 1 \) indicates that the text contains event descriptions, while \( D(t) = 0 \) indicates that it is not. By employing this systematic approach, we aim to ensure that the texts selected for further processing include relevant event descriptions and possess the structural characteristics necessary for effective code generation.

\subsection{Text-Code Cylic Training}

Predicting corresponding code from text input, and vice versa, lies at the heart of our generation's training, as shown in Figure \ref{fig:prompt2}. By iterating through this process using cyclic generation training, we harmonize text and code within a shared semantic space. Predicting code from text enhances the model's ability to maintain coherence and consistency throughout the generation process while predicting text from code helps bridge these distinct forms of semantic representation. This dual predictive approach not only reinforces the connection between text and code but also serves as a powerful strategy for enhancing language models' overall coherence and consistency. Below, I will provide a detailed introduction

\textbf{Text2Code} Based on the LLM model, designated as \( G \), to convert the input text \( t \) into corresponding structured code  \( c \). The generation process is guided by prompts, represented as \( p_{t2c} \), with specific instructions detailed in the Appendix A.2. This process can be expressed mathematically as:
\begin{equation}
c \sim G(\cdot | p_{t2c}, t),
\end{equation}

It is important to note that the instructions in the Appendix A.2 include a sample we crafted ourselves. This sample serves as a crucial reference for the base LLM model, enabling it to effectively comprehend our prompts and generate code in alignment with the target text. Without such a sample, the model would be unable to fulfill the task according to our expectations.

\begin{figure}[t]
    \includegraphics[height=7.8cm,width=7.7cm]{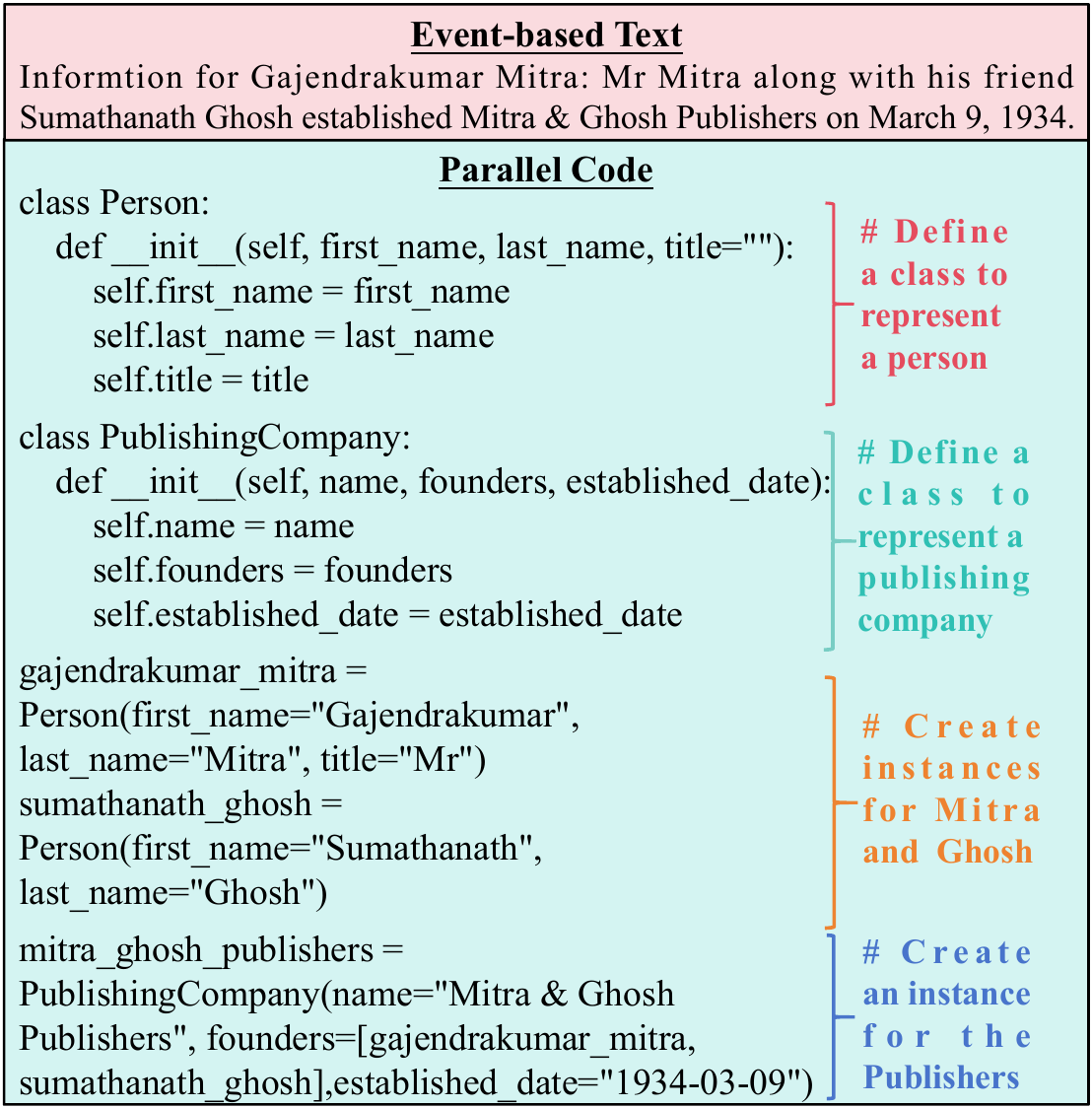}
    \centering
    \caption{An example of an original text and the generated code.}
    \label{fig:examples}
    \end{figure}

\textbf{Code2Text} Based on the base LLM model, we convert the input text into the corresponding structured code language. The generation process is guided by prompts, represented as \( p_{c2t} \), with specific instructions detailed in the Appendix A.3. This process can be expressed mathematically as:
\begin{equation}
t \sim G(\cdot | p_{c2t}, c),
\end{equation}
To provide a more intuitive understanding, Figure \ref{fig:examples} presents an example of converting event-driven text into code. In our case, we extract essential details about Gajendrakumar Mitra and Sumathanath Ghosh, the co-founders of "Mitra \& Ghosh Publishers," established on March 9, 1934. Leveraging this information, we define two classes: `Person`, which represents individual entities, and `PublishingCompany`, which encapsulates the characteristics of the publishing firm. Instances of these classes are created to accurately represent both the founders and the company, resulting in precise and high-quality code.

\subsection{Quality Assessment of Generated Code}

During iterative training, the code2text approach utilizes the original text as groundtruth, whereas the text2code process lacks a definitive "correct" code as groundtruth. This discrepancy hinders the implementation of iterative training. Therefore, a dynamic quality assessment of the code generated in each iteration is essential. This assessment can provide reliable feedback for text2code training, thus ensuring its effectiveness in subsequent model training.

We propose a quality evaluation method for dynamically generated code by integrating text and code. Given that the generated code is executable, we can represent entity names from the original text by invoking the properties of instantiated objects within the code. This process results in a reorganized text denoted as \( c_{\text{reorg}} \). To generate \( c_{\text{reorg}} \), We further instruct the LLM to generate this reorganized text based on predefined instructions \( p_{\text{reorg}}\) (detailed in the Appendix A.4)). The  process can be represented as:
\begin{equation}
c_{\text{reorg}} \sim G(\cdot | p_{\text{reorg}, t, c})
\end{equation}
We show an example in Figure \ref{fig:examples} to elucidate this process. The sentence \( c_{\text{reorg}}\) in Figure \ref{fig:examples} can be reorganized as:
\begin{center} 
\textit{
    \textcolor{blue}{\mbox{mitra\_ghosh\_publishers.founders[0].title}} 
    \textcolor{blue}{\mbox{mitra\_ghosh\_publishers.founders[0].first\_name}} 
    \\along with his friend 
    \textcolor{orange}{\mbox{mitra\_ghosh\_publishers.founders[1].first\_name} 
    \mbox{mitra\_ghosh\_publishers.founders[1].last\_name}} 
    \\established 
    \textcolor{cyan}{mitra\_ghosh\_publishers.name} \\on 
    \textcolor{magenta}{mitra\_ghosh\_publishers.established\_date}.
}
\end{center} 
Subsequently, we utilized a Python code interpreter to execute the corresponding code \( c \) as illustrated in Figure \ref{fig:examples}. Then, we replaced the placeholders with the output values obtained from running the code, resulting in the following generated content, denoted as \( c_{\text{reorg'}} \):

\begin{center} 
    \small 
    \textit{ 
        \textcolor{blue}{Mr} \textcolor{blue}{Mitra} \\along with his friend \\\textcolor{orange}{Sumathanath} \textcolor{orange}{Ghosh} \\established \textcolor{cyan}{Mitra \& Ghosh Publishers} \\on \textcolor{magenta}{March 9, 1934}.
    }
\end{center} 

If the quality of the generated code is high, the executed output will be more accurate, making \( c_{\text{reorg'}} \) more consistent with the original text \( t \). Conversely, if the quality of the parallel corpus is lower, the generated content may contain more errors and exhibit a reduced degree of similarity to the original text.

To systematically classify the quality of the generated output, we define a decision function \( D(S) \) based on the calculated similarity score:
\begin{equation}
D(S) = 
\begin{cases} 
1 & \text{if } S(c_{\text{reorg'}}, t) > T \\ 
0 & \text{if } S(c_{\text{reorg'}}, t) \leq T 
\end{cases}
\end{equation}
Let \( T \) be a predetermined threshold that distinguishes high-quality generated code from lower-quality outputs. The similarity score \( S \) is computed using the (ROUGE-L+ROUGE-1)/2, and if \( S \) falls below this threshold (set at 0.85 for our experiments), the generated data is deemed difficult and discarded. In contrast, if \( S \) meets or exceeds the threshold, it is considered to have the potential to generate high-quality parallel data and proceeds to the code-to-text phase. In our experiments, we set the number of iterative generations to 3; after each round of generation, all text outputs predict all corresponding code (including those discarded in the previous round). As the model improves its ability in iterative generation through training, we observe a higher proportion of successful text-to-code conversions. This demonstrates an enhancement in the text-code iterative generation capability, as well as the internal alignment of semantics and styles between the two languages. Ultimately, this allows the logical rigor of code to be transferred to natural language. 


\section{Experiment}
\subsection{Settings}


\textbf{Model} We assess the performance of several prominent base models, including Qwen2.5-7B\cite{yang2024qwen2}, Llama 3.1 Instruct-8B\cite{dubey2024llama}, and Ministra-8B-Instruct\cite{jiang2023mistral}. These models were selected based on their state-of-the-art capabilities in natural language processing tasks, providing a diverse range of architectures for our analysis.



\textbf{Dataset} We selected the wiki-40b-en dataset \cite{guo2020wiki} as source data for training in the consistency improvement of the model. This is a large-scale text dataset based on the English Wikipedia website. We validate our method on two tasks: text summarization using the CNN/Daily Mail dataset \cite{chen2016thorough}, which is a widely recognized benchmark, and question answering (QA) with HaluEval \cite{li2023halueval}, which assesses hallucination performance.


\textbf{Baseline} We selected the following baselines: 1) Base Model: Each of the selected models is evaluated in its original, instruct versions; 2) Naive Prompting: This approach involves adding straightforward, consistent instructions to guide the models' responses; 3) Supervised Fine-Tuning (SFT): Each base model undergoes SFT specific to the datasets used for the tasks; 4) SymbCOT \cite{xu2024faithful}: A recent method that reformulates chain-of-thought (CoT) reasoning as symbolic CoT to enhance consistency. The instructions for the 1) base and 2) prompt methods are shown in the Appendix A.5.


\textbf{Metric} For the summarization task, we utilize AlignScore \cite{zha2023alignscore} and consistency metric in UniEval\cite{zhong2022towards} to assess the phenomenon of inconsistent hallucination. In addition, we evaluate the summarization performance using its other three key metrics: Coherence, Fluency, and Relevance. For the QA task using the HaluEval dataset\cite{li2023halueval}, which is specifically designed to assess inconsistent hallucination, the actual performance of the QA task has not been evaluated. Instead, we use AlignScore\cite{zha2023alignscore} and Anah-V2\cite{gu2024anah} as metrics to investigate inconsistent hallucination, without focusing on task performance.


\textbf{Implementation Details} During the training in our method, we train each LLM with LLaMA-Factory on $4\times24$GB NVIDIA GeForce RTX 4090 GPUs. We use LoRA with a rank of 8 to accelerate training, with a learning rate of 1e-4. Unless otherwise specified, the training is conducted for 3 epochs, and the total batch size is 32. During the testing stage, since the models used are all instruct versions, we utilize their default chat templates and employ greedy decoding for text generation.

\subsection{Main Result}

Table \ref{tab:experiment1} presents the results of inconsistency hallucination for various baseline models across summarization and question-answering tasks, with optimal results highlighted in bold. Higher values are generally preferred except for the Anah-V2 metric, where lower values are favorable. The MIH-TCCT method achieves the highest performance across all metrics for the three models on the CNN/Daily Mail task. This trend is also seen in the HaluEval dataset, where optimal results are primarily associated with the SFT and MIH-TCCT methods. These findings indicate that MIH-TCCT effectively mitigates inconsistency hallucinations in generative models, ensuring logical output coherence.

Among the five evaluated methods, SymbCOT and MIH-TCCT enhance model consistency without task-specific adaptations, whereas SFT requires such adaptations. Overall, SFT and MIH-TCCT lead in performance, followed by SymbCOT and Prompt, with base methods showing the least favorable results. Notably, MIH-TCCT shows a significant advantage over SymbCOT and often surpasses the adapted SFT method. This analysis highlights MIH-TCCT’s effectiveness in reducing inconsistency hallucinations, reinforcing its superiority over other methods in enhancing the coherence of generative models.

Additionally, when comparing the impact of different baselines, the Llama 3.1-Instruct model consistently outperforms the other two models, particularly in base methods, achieving optimal results across both datasets. In terms of stability, models generally show consistent performance in summarization tasks, while QA tasks exhibit greater variability. Despite these differences, the MIH-TCCT method consistently demonstrates strong adaptability and stability across both task types.

\subsection{Model Ablation}

 Using the Llama 3.1-Instruct backbone, we evaluated all performance metrics across both the summarization and QA datasets by systematically removing specific components: the filter from Section 3.2, the quality assessment from Section 3.4, and the cyclic training discussed in Section 3.5, which was replaced with text generation code.
 
\begin{figure}[h!]
    \includegraphics[height=5.2cm,width=7.8cm]{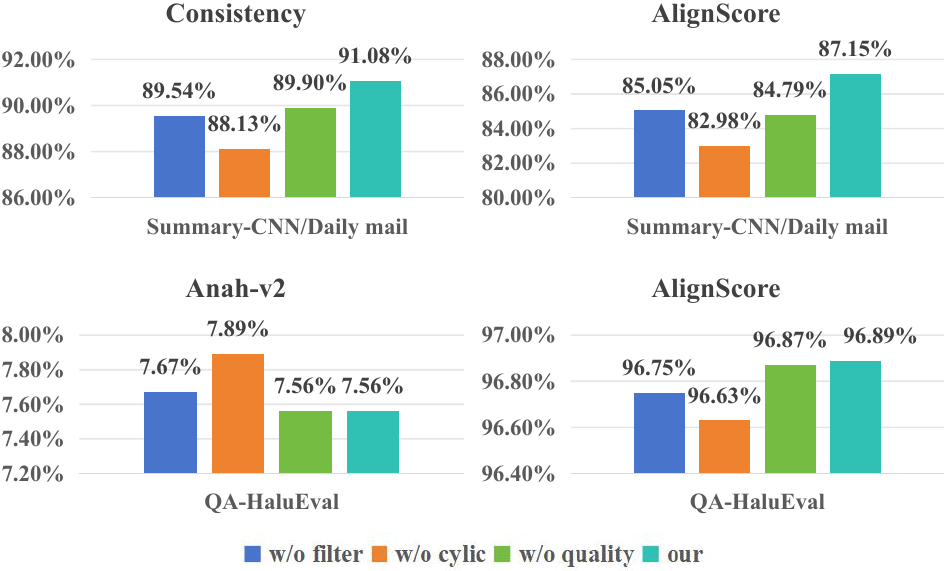}
    \centering
    \caption{The ablation experiment results based on the Llama 3.1-Instruct are presented for two types of tasks, displaying the inconsistency hallucination evaluation metrics across different tasks.}
    \label{fig:exp-ablation1}
    \end{figure}

 \textbf{Impact on hallucinations} As shown in Figure \ref{fig:exp-ablation1}, removing the filter, cyclic training, and quality assessment modules resulted in average declines of 0.97\%, 1.93\%, and 0.89\% across all consistency metrics for both tasks, with the cyclic training module having the most significant impact. The results confirm the effectiveness of various components within the MIH-TCCT framework.

\begin{table}[t]
    \centering
    \caption{The performance testing results for the inconsistency-type hallucination evaluation on the summarization task (CNN/Daily Mail dataset) and the QA task (HaluEval) are presented for three baseline models. Higher values are generally preferred, except for the Anah-V2 metric, where lower values are favored.}
    \setlength\tabcolsep{3.3pt}
    \small
    \begin{tabular}{@{}c@{}|c@{}|cc@{}|cc@{}}
        \hline
        \multirow{2}{*}{\textbf{Model}} & \multirow{2}{*}{\textbf{Method}} & \multicolumn{2}{c|}{\textbf{CNN/Daily mail}} & \multicolumn{2}{c}{\textbf{HaluEval}}  \\ 
        \cline{3-4} \cline{5-6} 
        & & \textbf{Consis} & \textbf{Align} & \textbf{Anah-v2} & \textbf{Align} \\ 
        \hline
        \multirow{7}{*}{Llama 3 }
        & Base            &86.40\%  &83.28\%& 12.48\% & 96.35\%  \\ 
        & Prompt          & 87.73\%&86.51\% &8.92\% & 96.17\%\\ 
        & SFT             &90.33\%&84.06\% &8.27\% &\textbf{98.52\%}  \\ 
        & SymbCOT       &86.80\%  & 84.28\% &11.66\% &96.58\% \\ 
        & MIH-TCCT   &\textbf{91.08\%}  &\textbf{87.15\%} & \textbf{7.56\%} & 96.82\%  \\ 
       \hline
        \multirow{7}{*}{Ministral} 
        & Base           &87.30\%&82.35\%&9.15\% & 97.16\%  \\ 
        & Prompt          & 88.02\%&83.12\% &8.12\%&97.11\%\\ 
        & SFT             &90.10\%&83.54\%&8.11\%&\textbf{98.61\%}\\ 
        & SymbCOT        &88.26\% &82.10\% &8.60\%&97.14\%  \\
        & MIH-TCCT    &\textbf{90.98\%} & \textbf{86.21\%} &\textbf{7.85\%}& 97.50\% \\ 
         \hline
        \multirow{7}{*}{Qwen} 
        & Base            &83.76\% &82.56\%& 10.75\% & 95.05\%  \\ 
        & Prompt          & 84.51\% &84.83\% & 9.98\%& 95.14\% \\ 
        & SFT            &89.12\% &81.13\% &\textbf{8.14\%} &\textbf{98.39\%}   \\ 
         & SymbCOT          &84.45\% &81.84\%&11.38\% & 95.35\%   \\ 
        & MIH-TCCT    &\textbf{90.70\%}&\textbf{85.53\%} & 9.22\% &96.67\%  \\ 
        \hline
    \end{tabular}
    \label{tab:experiment1}
\end{table}

\begin{figure}[h!]
\includegraphics[height=5.2cm,width=7.8cm]{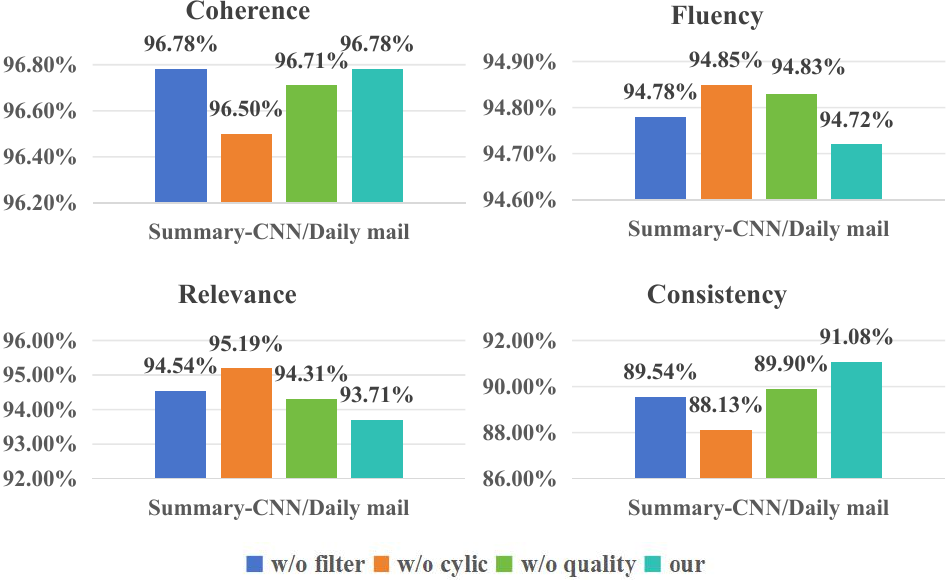}
\centering
\caption{The ablation experiment results for the summarization task using the Llama 3.1-Instruct backbone are presented. The QA task from the HaluEval dataset is designed to evaluate hallucinations but lacks performance metrics for QA itself. Additionally, the consistency metric serves both as a measure of consistency and an evaluation metric for the summarization task.}
\label{fig:exp-ablation2}
\end{figure}

\textbf{Impact on the summarization task} As shown in Figure \ref{fig:exp-ablation2},the average reductions in the four metrics were 0.16\%, 0.41\%, and 0.14\%, respectively. The effect on summarization performance was minimal, which may be attributed to our framework reducing the need for downstream task adaptation, thus making the performance impact on downstream tasks negligible.

\begin{table}[t]
    \centering
    \caption{The performance on the summarization task (CNN/Daily Mail dataset) include four metrics: coherence, consistency, fluency, and relevance. Due to space constraints, these metrics are abbreviated in the table as Cohere, Consis, Fluency, and Releva. Additionally, the consistency metric serves both as a measure of hallucination and an evaluation metric for the summarization task.}
    \setlength\tabcolsep{3.5pt}
    \small
    \begin{tabular}{@{}c@{}|c@{}|cccc@{}}
        \hline
        \multirow{2}{*}{\textbf{Model}} & \multirow{2}{*}{\textbf{Method}} & \multicolumn{4}{c}{\textbf{CNN/Daily mail}}  \\ 
        \cline{3-4} \cline{5-6} 
        &  & \textbf{Cohere}& \textbf{Consis} &\textbf{Fluency} & \textbf{Releva} \\ 
        \hline
        \multirow{7}{*}{Llama 3}
        & Base           & 96.40\% &86.40\%& 94.84\% & \textbf{95.61\%} \\ 
        & Prompt         & \textbf{96.82\%} &87.73\%& \textbf{94.88\%} & 95.56\% \\ 
        & SFT            & 90.56\% &90.33\%& 93.21\% & 81.53\%  \\ 
        & SymbCOT       & 96.52\%&86.80\% & 94.86\% & 95.47\%  \\ 
        & MIH-TCCT   & 96.78\%&\textbf{91.08\%} &	94.72\% &93.71\%   \\ 
       \hline
        \multirow{7}{*}{Ministral} 
        & Base           & 95.62\%&87.30\% & 94.78\% &  \textbf{94.20\%}  \\ 
        & Prompt          &  \textbf{95.84\%} &88.02\%& 94.77\% & 93.96\% \\ 
        & SFT             &  90.74\%&90.10\%&92.99\% &82.51\% \\  
        & SymbCOT       & 94.95\% &88.26\&& \textbf{94.81\%} & 92.25\%\\         
        & MIH-TCCT    &95.75\%&\textbf{90.98\%}&	94.44\%&91.70\%\\ 
         \hline
        \multirow{7}{*}{Qwen} 
        & Base            & 95.11\%&83.76\% & 94.63\% & \textbf{94.60\%} \\ 
        & Prompt          & 95.36\%&84.51\%&	94.73\%&94.27\% \\ 
        & SFT           & 89.41\%&89.12\% & 93.33\% & 81.79\% \\ 
        & SymbCOT          & 94.65\% &84.45\%& \textbf{95.00\%} & 93.68\% \\ 
        & MIH-TCCT  &  \textbf{95.52\%}&\textbf{90.70\%} &94.60\%  &90.41\%   \\ 
        \hline
    \end{tabular}
    \label{tab:experiment11}
\end{table}

\subsection{Analysis and Discussion}
\subsubsection{Impact on Task Performance}

To assess whether our approach affects the overall performance of the task while reducing inconsistent hallucinations, we examine the metrics of coherence, consistency, fluency, and relevance in the summarization task, with detailed results presented in Table \ref{tab:experiment11}. Optimal results are highlighted in bold. Table \ref{tab:experiment11} indicates that our MIH-TCCT method effectively maintains task performance across most metrics without significant decline. In contrast, while the SFT method shows consistent competitive results, it experiences notable declines in relevance and coherence, indicating instability. This emphasizes the advantage of MIH-TCCT in providing balanced performance across all metrics while effectively reducing inconsistencies.

\subsubsection{Analysis of the Cyclic Training}

The text-code cyclic training not only enhances the alignment capabilities of the LLM in both text and programming spaces but also progressively improves its generative abilities. As illustrated in Figure 2, the similarity scores for both the reconstructed text and the original text increase with the number of training epochs, with a maximum difference of 5.23\%. Simultaneously, the proportion of data meeting the quality standards steadily rises, achieving an enhancement of up to 14.55\%. Additionally, by the third training round, the upward trend stabilizes, leading us to set the training duration at three epochs.

\subsubsection{Impact of Homogeneous Data}

\begin{figure}[t]
    \includegraphics[height=4.1cm,width=7.7cm]{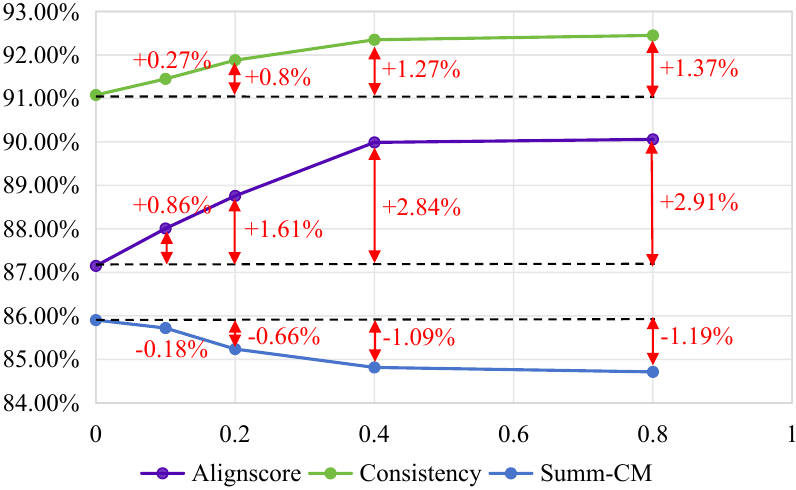}
    \centering
    \caption{Performance of the model based on the Llama3.1-Instruct architecture across varying proportions of mixed homogeneous data on the CNN-Daily Mail dataset.}
    \label{fig:exp-propo}
    \end{figure}

The main experiments were conducted on the wiki-40b-en dataset, with validation on heterogeneous data demonstrating the model's logical consistency and generalization capability. To further assess performance, we examined the impact of increasing homogeneous data during training. Figure \ref{fig:exp-propo} presents the results from varying proportions of mixed homogeneous data using the Llama3.1-Instruct base model on the CNN-Daily Mail dataset for summarization tasks. The x-axis is on an exponentially increasing scale, while the vertical axis shows the model's performance, including hallucination metrics via AlignScore, Consistency, and the multiplicative results of three other summarization metrics, labeled as Summ-CM. Point 0.0 reflects results from heterogeneous data alone, with red lines indicating changes relative to these results.

As illustrated in Figure \ref{fig:exp-propo} (details in Table \ref{tab:experiment22} of the Appendix), our findings suggest that a small proportion of mixed homogeneous data is sufficient for aligning the semantic spaces of code and data, resulting in significant improvements in consistency metrics. While increasing the homogeneous data enhances the logical consistency of the output, the overall performance metrics for summarization tasks decline, stabilizing at a parameter setting of 0.4. Consequently, we selected 0.4 for further experiments, with results presented in Table \ref{tab:model_performance}, and the results of adding heterogeneous data recorded as AdvTrain. The AdvTrain method shows significant performance improvements over MIH-TCCT across nearly all four hallucination evaluation metrics, particularly in the summarization task relative to the QA task.

\begin{table}[t]
    \centering
    \caption{Building on MIH-TCCT, we increased the proportion of homogeneous data by 40\% (denoted as AdvTrain in the table below) and trained the model using the Llama3.1 base model. We evaluated whether this adjustment would further mitigate the model's hallucination across summarization and QA tasks.}
    \setlength\tabcolsep{3.2pt}
    \small
    \begin{tabular}{@{}c@{}|c@{}|cc@{}|cc@{}}
        \hline
        \multirow{2}{*}{\textbf{Model}} & \multirow{2}{*}{\textbf{Method}} & \multicolumn{2}{c|}{\textbf{CNN/Daily mail}} &\multicolumn{2}{c}{\textbf{HaluEval}} \\
        \cline{3-6} 
         & & \textbf{Consis} & \textbf{Align} & \textbf{Anah-v2} & \textbf{Align} \\ 
        \hline
         \multirow{2}{*}{Llama 3.1 } & AdvTrain   & \textbf{92.35\%}& \textbf{89.99\%} &\textbf{7.38\%} & \textbf{96.89\%}    \\
        & MIH-TCCT   & 91.08\%  & 87.15\%    & 7.56\%    & 96.82\%    \\ 
        \hline
       \multirow{2}{*}{Ministral} & AdvTrain    & \textbf{91.28\%}  & \textbf{86.77\%}& 7.99\%  & 97.39\%    \\
        & MIH-TCCT &90.98\%&86.21\%&\textbf{7.85\%}&\textbf{97.50\%} \\ 
        \hline
         \multirow{2}{*}{Qwen} & AdvTrain   & \textbf{91.64\%} & \textbf{87.15\%} &\textbf{9.00\%}& \textbf{96.74\%}    \\
        & MIH-TCCT   & 90.70\%  & 85.53\%  &9.22\% & 96.67\%\\ 
        \hline
    \end{tabular}
    \label{tab:model_performance}
\end{table}

\subsubsection{Analysis of the Text Filter Module}

We randomly selected 10,000 entries from the wiki-40b-en dataset to evaluate the text filter. As shown in Figure \ref{fig:exp1} in the Appendix, only 53.74\% of the texts describe events, underscoring the need for an event-type text filter. To assess the filter's performance, we extracted 100 samples each of event-type and non-event-type texts. These samples were annotated by three doctoral students, with final classifications determined through a voting process. The confusion matrix in Figure \ref{fig:exp1} shows a high true positive rate (0.93) and low false negative rate (0.07) for event-type texts, as well as a strong true negative rate (0.92) and low false positive rate (0.08) for non-event-type texts. This confirms the model's effectiveness in distinguishing relevant content from irrelevant content.

   \begin{figure}[t]
    \includegraphics[height=3.8cm,width=7.8cm]{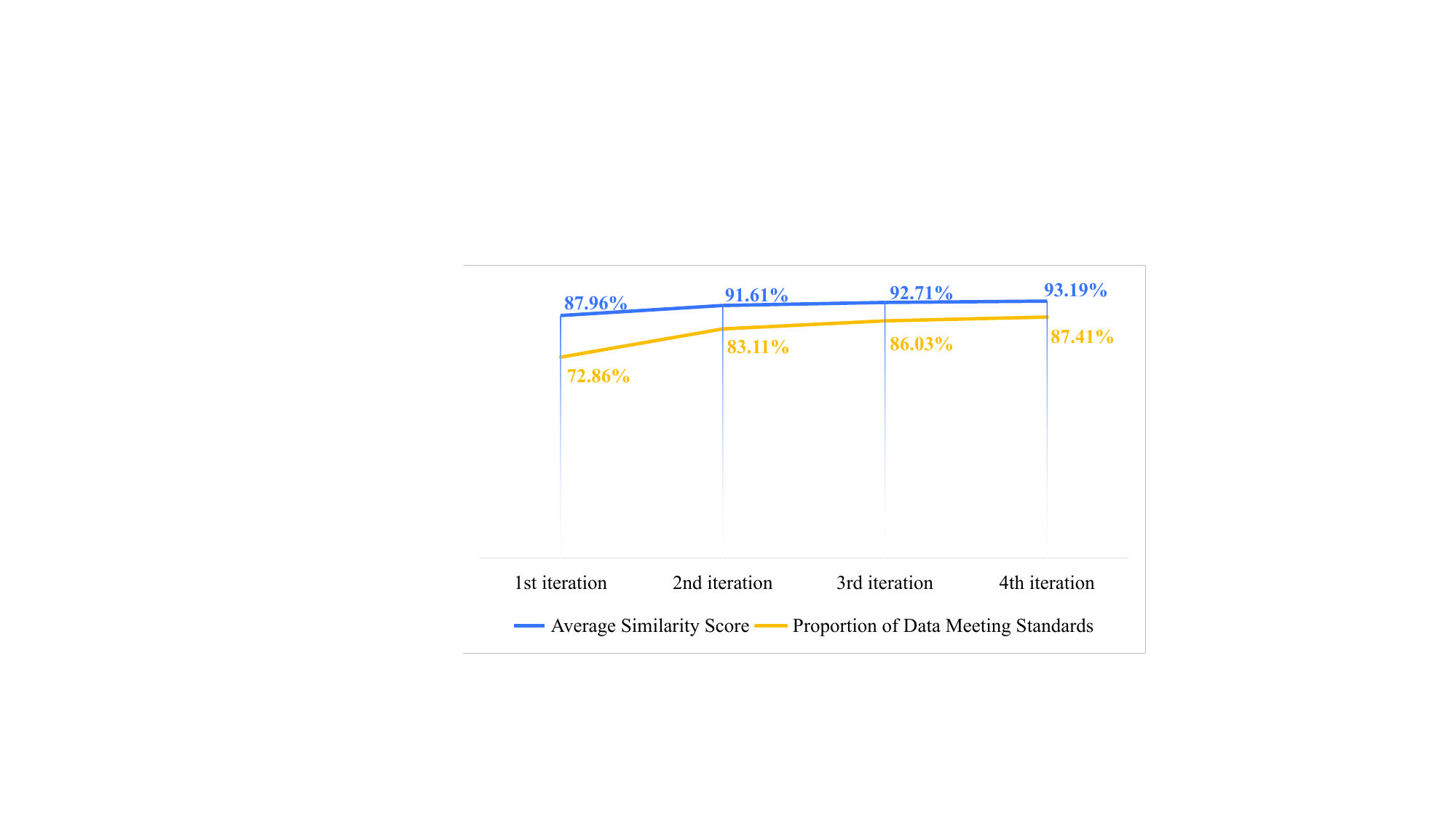}
    \centering
    \caption{The trend of changes in the similarity score and the proportion of qualifying data in the quality assessment module as the number of cyclic training iterations increases.}
    \label{fig:exp1}
    \end{figure}
\section{Conclusion}

We introduce MIH-TCCT, a novel framework that enhances the logical consistency of LLM outputs by cyclically generating event-based text and corresponding code. This approach effectively transfers the logical rigor of code to natural language, addressing significant challenges related to inconsistent hallucinations. Notably, our framework demonstrates exceptional generalizability, meaning that it can alleviate issues of hallucination across various contexts without adapting to specific downstream tasks. These findings contribute to advancing the capabilities of LLMs, paving the way for more reliable and logically consistent AI systems.

\section*{Impact Statement}

This work introduces MIH-TCCT, a novel framework that enhances the logical consistency of LLM outputs by cyclically generating event-based text and corresponding code. This approach effectively transfers the logical rigor of code to natural language, addressing significant challenges related to inconsistent hallucinations. The datasets and codes used in this study are built based on public medical datasets and will be publicly available to facilitate further research. We have not identified any potential negative ethical consequences requiring
further consideration.

\nocite{langley00}

\bibliography{example_paper}

\newpage
\appendix
\onecolumn

\section{Introduction to the Prompt Used}

\subsection{The Prompt for Filtering Texts}


\small 
\textit{ 
\textcolor{blue}{Your task is to determine whether a text is suitable for description in code (e.g., modeling a relationship using class-based instances, modeling a process using a function), and return “yes” if it is suitable, or “no” otherwise. Generally speaking, an organized text that describes a series of events and processes is suitable for code description, while a text that mainly expresses emotions or illogicality is not suitable for code description, the following is an examples:}\\
\textcolor{blue}{Input :} The Hullaballoos were created in August 1964, but had been working in the UK for over three years under the name of Ricky Knight and The Crusaders. They They were not named after the American television programme Hullabaloo. Their name came from the city of Hull, England, whence they hailed. \\
\textcolor{blue}{Output : }yes\\
\textcolor{blue}{Please decide whether the following text is suitable for code description according to the above requirements:}\\
FK Drezga is founded on 1972, as a team of Piperi region near Podgorica. At their first seasons, FK Drezga played in Fourth League - Central region (lowest rank in SFR Yugoslavia). Club was dissaloved at the end of seventies.
}

\subsection{The Prompt for Predicting Code from Text}

The prompt provided to the trained base model for cyclic training in predicting code from text.

\small 
\textit{ 
\textcolor{blue}{Please use a code snippet to encapsulate the given text into a structured format using classes and instances in Python. Here is an example for you.}\\
\textcolor{blue}{Input:}for Gajendrakumar Mitra: Mr Mitra along with his friend Sumathanath Ghosh established Mitra \& Ghosh Publishers on March 9, 1934.\\
\textcolor{blue}{output:}\\
\textcolor{red}{\#Define a class to represent a person}\\
class Person:\\
   \hspace*{1.5em} def \underline{\hspace{0.3cm}}init\underline{\hspace{0.3cm}}(self, first\underline{\hspace{0.15cm}}name, last\underline{\hspace{0.15cm}}name, title=""):\\
        \hspace*{3.1em}self.first\underline{\hspace{0.15cm}}name = first\underline{\hspace{0.15cm}}name\\
        \hspace*{3.1em}self.last\underline{\hspace{0.15cm}}name = last\underline{\hspace{0.15cm}}name\\
        \hspace*{3.1em}self.title = title\\
\textcolor{red}{\#Define a class to represent a publishing company}\\
class PublishingCompany:\\
    \hspace*{1.5em}def \underline{\hspace{0.3cm}}init\underline{\hspace{0.3cm}}(self, name, founders, established\underline{\hspace{0.15cm}}date):\\
        \hspace*{3.1em}self.name = name\\
        \hspace*{3.1em}self.founders = founders\\
        \hspace*{3.1em}self.established\underline{\hspace{0.15cm}}date = established\underline{\hspace{0.15cm}}date\\
\textcolor{red}{\# Create instances for Gajendrakumar Mitra and Sumathanath Ghosh}\\
gajendrakumar\underline{\hspace{0.15cm}}mitra = Person(first\underline{\hspace{0.15cm}}name="Gajendrakumar", last\underline{\hspace{0.15cm}}name="Mitra", title="Mr")\\
sumathanath\underline{\hspace{0.15cm}}ghosh = Person(first\underline{\hspace{0.15cm}}name="Sumathanath", last\underline{\hspace{0.15cm}}name="Ghosh")\\
\textcolor{red}{\# Create an instance for the publishing company Mitra \& Ghosh Publishers}\\
mitra\underline{\hspace{0.15cm}}ghosh\underline{\hspace{0.15cm}}publishers = PublishingCompany(
    name="Mitra \& Ghosh",
    Publishers founders=[gajendrakumar\underline{\hspace{0.15cm}}mitra,sumathanath\underline{\hspace{0.15cm}}ghosh],
    established\underline{\hspace{0.15cm}}date="1934-03-09")\\
\textcolor{blue}{Please complete a encapsulation of the following text based on the above requirements and particles.}\\
Bacon was born in Ipswich and lived in Great Yarmouth as a child. Bacon attended Yarmouth Art School from 1917-1923,where she won a scholarship in 1917 and by 1921 passed the Board of Education drawing examinations at the earliest age possible. She studied at the Norwich School of Art and then at the Royal College of Art in London, obtaining her diploma in 1927.
}


\subsection{The Prompt for Predicting Text from Code}

The prompt provided to the trained base model for cyclic training in predicting text from code.

\small 
\textit{ 
\textcolor{blue}{Here is a code snippet which encapsulates a original text into a structured format using classes and instances in Python. You are going to predict the original text after reading the code}\\
\textcolor{red}{\#Define a class to represent a person}\\
class Person:\\
   \hspace*{1.5em} def \underline{\hspace{0.3cm}}init\underline{\hspace{0.3cm}}(self, first\underline{\hspace{0.15cm}}name, last\underline{\hspace{0.15cm}}name, title=""):\\
        \hspace*{3.1em}self.first\underline{\hspace{0.15cm}}name = first\underline{\hspace{0.15cm}}name\\
        \hspace*{3.1em}self.last\underline{\hspace{0.15cm}}name = last\underline{\hspace{0.15cm}}name\\
        \hspace*{3.1em}self.title = title\\
\textcolor{red}{\#Define a class to represent a publishing company}\\
class PublishingCompany:\\
    \hspace*{1.5em}def \underline{\hspace{0.3cm}}init\underline{\hspace{0.3cm}}(self, name, founders, established\underline{\hspace{0.15cm}}date):\\
        \hspace*{3.1em}self.name = name\\
        \hspace*{3.1em}self.founders = founders\\
        \hspace*{3.1em}self.established\underline{\hspace{0.15cm}}date = established\underline{\hspace{0.15cm}}date\\
\textcolor{red}{\# Create instances for Gajendrakumar Mitra and Sumathanath Ghosh}\\
gajendrakumar\underline{\hspace{0.15cm}}mitra = Person(first\underline{\hspace{0.15cm}}name="Gajendrakumar", last\underline{\hspace{0.15cm}}name="Mitra", title="Mr")\\
sumathanath\underline{\hspace{0.15cm}}ghosh = Person(first\underline{\hspace{0.15cm}}name="Sumathanath", last\underline{\hspace{0.15cm}}name="Ghosh")\\
\textcolor{red}{\# Create an instance for the publishing company Mitra \& Ghosh Publishers}\\
mitra\underline{\hspace{0.15cm}}ghosh\underline{\hspace{0.15cm}}publishers = PublishingCompany(
    name="Mitra \& Ghosh",
    Publishers founders=[gajendrakumar\underline{\hspace{0.15cm}}mitra,sumathanath\underline{\hspace{0.15cm}}ghosh],
    established\underline{\hspace{0.15cm}}date="1934-03-09")\\
}

    

\subsection{The Prompt for Predicting Reorganized Text }

The prompt provided to the trained base model to predict reorganized text based on generated code.

\small 
\textit{ 
\textcolor{blue}{Please further utilize the original text and the generated code data to represent the entities in the original text with code, forming a new mixed text. Below is an example.}\\
text=f"{mitra\underline{\hspace{0.15cm}}ghosh\underline{\hspace{0.15cm}}publishers.founders[0].title} \\{mitra\underline{\hspace{0.15cm}}ghosh\underline{\hspace{0.15cm}}publishers.founders[0].first\underline{\hspace{0.15cm}}name} \\{mitra\underline{\hspace{0.15cm}}ghosh\underline{\hspace{0.15cm}}publishers.founders[0].last\underline{\hspace{0.15cm}}name} along with his friend \\{mitra\underline{\hspace{0.15cm}}ghosh\underline{\hspace{0.15cm}}publishers.founders[1].first\underline{\hspace{0.15cm}}name} \underline{}{mitra\underline{\hspace{0.15cm}}ghosh\underline{\hspace{0.15cm}}publishers.founders[1].last\underline{\hspace{0.15cm}}name} established \\{mitra\underline{\hspace{0.15cm}}ghosh\underline{\hspace{0.15cm}}publishers.name} on \\{mitra\underline{\hspace{0.15cm}}ghosh\underline{\hspace{0.15cm}}publishers.established\underline{\hspace{0.15cm}}date}.}

\subsection{The Prompt for the Base and Prompt Versions Used for Model Testing}

\begin{table*}[h]
    \centering
    \caption{For the summary and QA datasets, show the instructions for the base and prompt versions of the three model bases.}
    \setlength\tabcolsep{3.5pt}
    \small
    \begin{tabular}{p{3cm}|p{5cm}|p{5cm}@{}} 
        \hline
        \textbf{Datasets} & \textbf{Base} & \textbf{Prompt} \\
        \hline
        CNN/Daily Mail & Write a short summary of the following news. & Write a short summary of the following news. Attention should be paid to the consistency of the abstract with the original text to avoid generating content with hallucination. \\
        \hline
        HaluEval & You are a question answerer. You should answer the questions directly based on the given reference without adding any prefixes or suffixes, and without analyzing the answers. After answering the question, do not say anything else. Reference document: {} Please answer the question based on the above reference: {} & You are a question answerer. You should answer the questions directly based on the given reference without adding any prefixes or suffixes, and without analyzing the answers. After answering the question, do not say anything else. Please do not output content that is inconsistent with the context, and avoid giving irrelevant or contradictory answers. Reference document: {} Please answer the question based on the above reference: {} \\
        \hline
    \end{tabular}
    \label{tab:prompt11}
\end{table*}

\section{Supplement to the Experimental Section}

\begin{figure}[htbp!]
    \includegraphics[height=3.2cm,width=8cm]{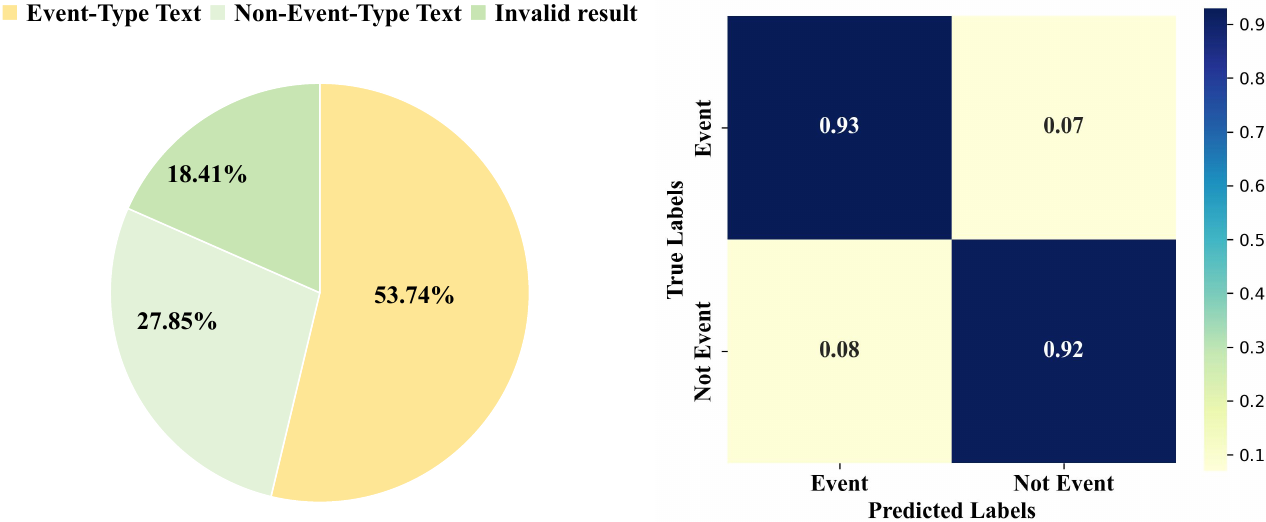}
    \centering
    \caption{The left figure illustrates the proportions of event-type text, non-event-type text, and invalid results after applying the event-type text filter. The right figure presents the confusion matrix, which captures the classification performance in distinguishing between event-type and non-event-type texts.}
    \label{fig:exp1}
    \end{figure}

\begin{table}[t]
    \centering
    \caption{The performance of the model based on the Llama3.1-Instruct architecture across varying proportions of mixed homogeneous data using the CNN-Daily Mail dataset. The performance assessment incorporates three crucial metrics: Coherence, Fluency, and Relevance. The "Summ-CM" score is calculated by multiplying these three metrics together.}
    \begin{tabular}{c|cc|ccc|c}
        \hline
        \multirow{2}{*}{\textbf{Ratio}}  & \multicolumn{6}{c}{\textbf{Summary-CNN/Daily mail}}  \\ 
        \cline{2-4} \cline{5-7} 
        &  \textbf{Consistency}& \textbf{AlignScore} & \textbf{Coherence} & \textbf{Fluency} & \textbf{Relevance} & \textbf{Summ-CM} \\
       \hline
       0.0  & 91.08\%     & 87.15\%     & 96.78\%   & 94.72\% & 93.71\%   & 85.90\% \\
        0.1   & 91.45\%     & 88.01\%     & 96.76\%   & 94.78\% & 93.47\%   & 85.72\% \\
        0.2   & 91.88\%     & 88.76\%     & 96.74\%   & 94.64\% & 93.10\%   & 85.24\% \\
        0.4   & 92.35\%     & 89.99\%     & 96.84\%   & 94.50\% & 92.68\%   & 84.81\% \\
        0.8   & 92.45\%     & 90.06\%     & 96.90\%   & 94.46\% & 92.55\%   & 84.71\% \\
        \hline
    \end{tabular}
    \label{tab:experiment22}
\end{table}




\end{document}